\pdfoutput=1

\documentclass[11pt]{article}

\usepackage[]{emnlp2021}

\usepackage{times}
\usepackage{latexsym}
\usepackage{makecell}
\usepackage{pifont}
\newcommand{\cmark}{\ding{51}}%
\newcommand{\xmark}{\ding{55}}%

\usepackage[T1]{fontenc}

\usepackage[utf8]{inputenc}

\usepackage{microtype}

\usepackage{graphicx}
\usepackage{subfigure}

\title{What Would it Take to get Biomedical QA Systems into Practice?}

\author{Gregory Kell \and Iain J. Marshall \\
  King's College London\\
  \texttt{\{name.surname\}@kcl.ac.uk} \\\And
  Byron C. Wallace \\
  Northeastern University \\ 
  \texttt{b.wallace@northeastern.edu} \\\AND
  Andr\'e Jaun \\
  Metadvice \\
  \texttt{ajaun@metadvice.com} \\}

\begin{document}
\maketitle

\begin{abstract}
Medical question answering (QA) systems have the potential to answer clinicians' uncertainties about treatment and diagnosis on-demand, informed by the latest evidence.  
However, despite the significant progress in general QA made by the NLP community, medical QA systems are still not widely used in clinical environments.
One likely reason for this is that clinicians may not readily trust QA system outputs, in part because transparency,  trustworthiness, and provenance have not been key considerations in the design of such models. In this paper we discuss a set of criteria that, if met, we argue would likely increase the utility of biomedical QA systems, which may in turn lead to adoption of such systems in practice.
We assess existing models, tasks, and datasets with respect to these criteria, highlighting shortcomings of previously proposed approaches and pointing toward what might be more usable QA systems.
\end{abstract}

\section{Introduction}

During consultations in primary care, clinicians generate at least one question for every two patients \cite{del_fiol_clinical_2014}. Nonetheless, clinicians look for answers to only half of the questions due to time constraints and the belief that answers to certain questions do not exist \cite{del_fiol_clinical_2014}, despite the plethora of available evidence \cite{bastian_seventy-five_2010}. 
When clinicians do search for answers, they usually spend fewer than 
three minutes per question doing so \cite{del_fiol_clinical_2014,hoogendam_answers_2008}. 

Our focus in this paper is on questions pertaining to patient care decisions, for example seeking guidance about diagnosis or treatment. Ideally, clinicians would search for answers to such questions with reference to high-quality studies and up-to-date evidence syntheses,  typically indexed in medical databases such as PubMed\footnote{\url{https://pubmed.ncbi.nlm.nih.gov}} and the Cochrane Library.\footnote{\url{https://www.cochranelibrary.com}}
This practice of emphasizing use of rigorous empirical evidence is known as \emph{evidence-based medicine} (EBM). 
Under this framework, evidence compiled from \emph{all relevant high-quality research} (in the form of, e.g., systematic reviews and rigorously produced clinical guidelines) is preferred to individual studies or expert opinion \cite{ebell_strength_2004, guyatt_grade_2008, alper_ebhc_2016}. 
Unfortunately, searching existing sources for relevant, high-quality information is onerous.
Due to the time constraints imposed on clinicians, this leads to widespread reliance on general information sources such as Google \cite{hider_information-seeking_2009}.
However, while simple to use, general-purpose search engines rank results 
according to criteria not directly aligned with EBM principles such as rigour, comprehensiveness, and reliability \cite{hider_information-seeking_2009}.



Aside from internet search, clinicians often engage in informal discussions about decisions with colleagues in what are sometimes referred to as ``curbside consultations'' \cite{papermaster_common_2017,papermaster_exploring_2020,oleary_information-seeking_2012}.
It is common for practitioners to engage in at least one such discussion per week for practical reasons, including convenience, 
or an urgent need for information \cite{smith_what_1996}.
 These inform the ``mindlines'' that clinicians acquire over their careers (i.e., mental models of medicine) and that are also based on other sources including guideline documents, training, background reading, and experience \cite{gabbay_mindlines_2016}.
However, the information exchanged in informal consultations may be inaccurate, incomplete, and lead to practice influenced more by expert opinion than the scientific literature \cite{papermaster_common_2017}.

Medical question answering (QA) systems have the potential to address these issues by answering clinicians' questions in real-time on the basis of the latest evidence.
This has motivated development of QA systems and associated medical QA datasets used to train them.  For example, BioASQ \cite{tsatsaronis_overview_2015} and PubMedQA \cite{jin_pubmedqa_2019} have been created to train and evaluate systems that answer clinicians' questions based on medical research literature, while emrQA \cite{pampari_emrqa_2018}, emrKBQA \cite{Raghavan2021emrKBQAAC} and why-QA \cite{fan-2019-annotating} were constructed using queries concerning patient data from electronic health records (EHRs).
MEDIQA-QA \cite{ben-abacha-etal-2019-overview} and LiveQA-Medical  \cite{Abacha2017OverviewOT} are  datasets designed for systems that answer consumer (patient) queries.  MEDIQA-AnS \cite{savery_question-driven_2020} accompanies the answers from MEDIQA-QA with summaries that consumers would understand more easily.  
Systems for QA over EHRs aim to answer questions about the medical history or prior care of individual patients. 
By contrast, our focus here is on systems that can provide general evidence-based guidance in response to queries; we therefore omit emrQA, emrKBQA and why-QA from our discussion. 

Existing biomedical QA systems that answer questions with reference to the medical literature typically provide answers in the form of yes/no, factoids, lists, and/or definitions \cite{sarrouti_sembionlqa_2020, ben_abacha_means_2015, cao_askhermes_2011, zahid_cliniqa_2018, 10.1016/j.jbi.2007.03.002} without supplying justifications, e.g., source journals, extracted text snippets, and/or associated statistics. However, this answer format does not readily translate into clinical practice.

Take, for example, the question ``Which antibiotic should I use for urinary tract infections?''. A factoid-based QA system might (reasonably) return the answer ``trimethoprim 200mg''. However, a ``correct'' answer is not sufficient to translate into clinical use. An answer here is only as reliable as the source from which it was extracted. The source therefore needs to be judiciously chosen, and presented transparently.
Furthermore, in this example, the knowledge of the best treatment requires information about the patients' age and any additional health problems (for instance, dosing may vary in children, or where someone has impaired kidney function). The optimal treatment might vary by location, reflecting local or individual bacterial resistance patterns (which frequently change over time), or vary depending on the cost of drug acquisition or availability. 
A factoid answer does not allow the possibility of changing practice, or providing critical information which is not a direct response to the narrow question asked (perhaps an antibiotic is not always needed). 
These issues both need to be considered in producing an answer, and need to be \emph{seen} to have been considered by the clinician before s/he can feel confident in following the recommendation.

In this context, reliability is multi-faceted. 
For an answer to be reliable it must have been extracted from a trustworthy source, accurately transcribed, and relevant to the clinical context (was the dosing information extracted for the correct clinical condition?). 
It should also be locally applicable, and recent. In this example, a methodologically sound national clinical guideline is likely to be highly dependable, whereas a journal editorial or case study giving one expert's idiosyncratic opinion might be safely ignored. 
A question-answering system which does not understand the difference is not likely to be useful.


We argue that the deployment of EBM-guided QA systems---by which we mean those intended to provide answers to clinical questions based on published evidence---in clinical practices is contingent on the outputs being reliable and actionable. 
Clinicians should be able to trust that the most robust evidence was retrieved, and that conflicting evidence was handled appropriately. 
Uncertainties associated with answers should be communicated to the clinician.


\begin{figure}[ht]
\centering
\includegraphics[scale=0.08]{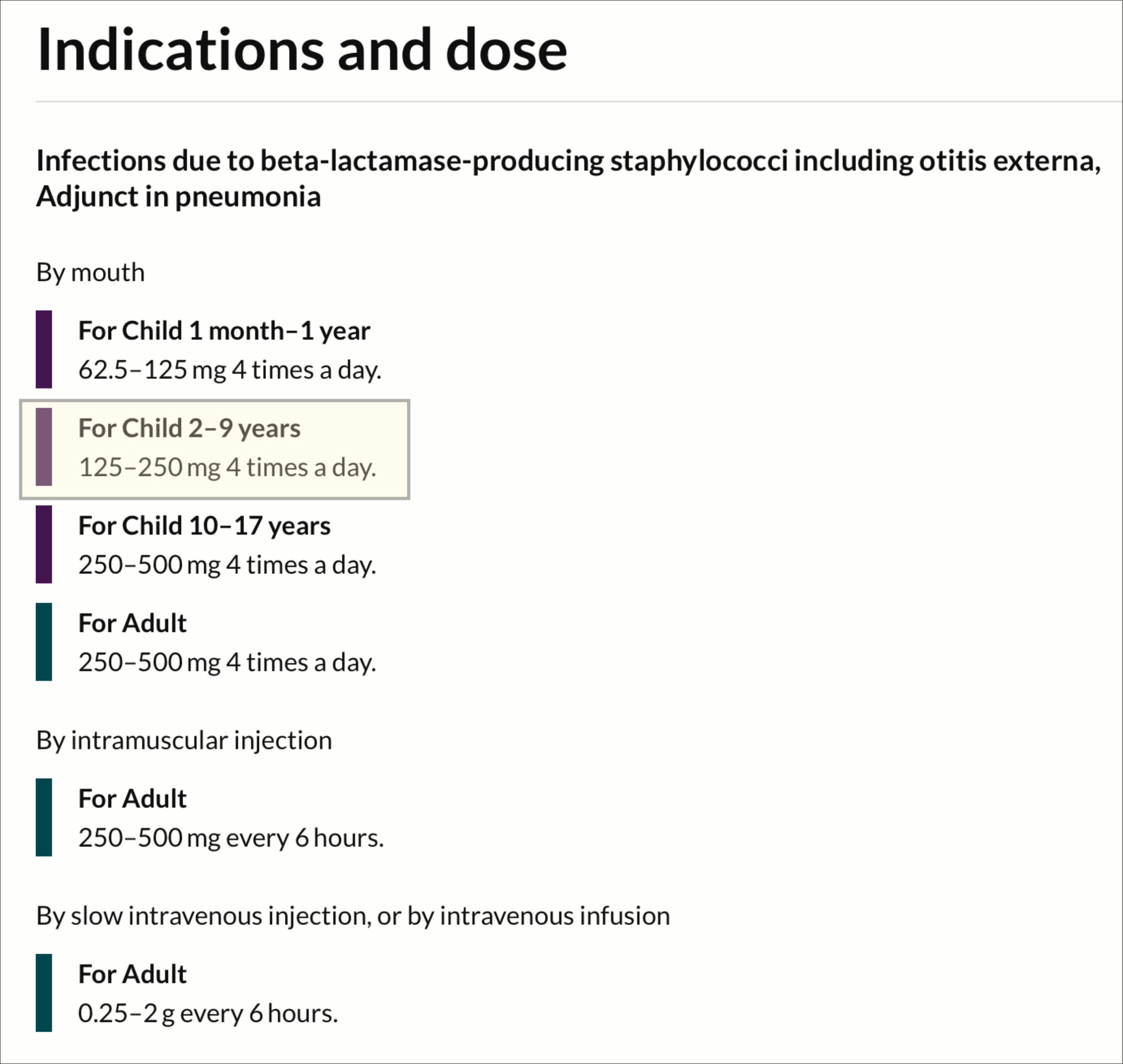}
\caption{Yellow box shows text snippet used to answer ``what dose of flucloxacillin should I prescribe for a 5 year old child?''\footnote{\url{https://bnf.nice.org.uk/drug/flucloxacillin.htm}}.}
\label{fig:rationale_extracted_text}
\end{figure}


\begin{figure*}[!ht]

  \centering
  \includegraphics[scale=0.25]{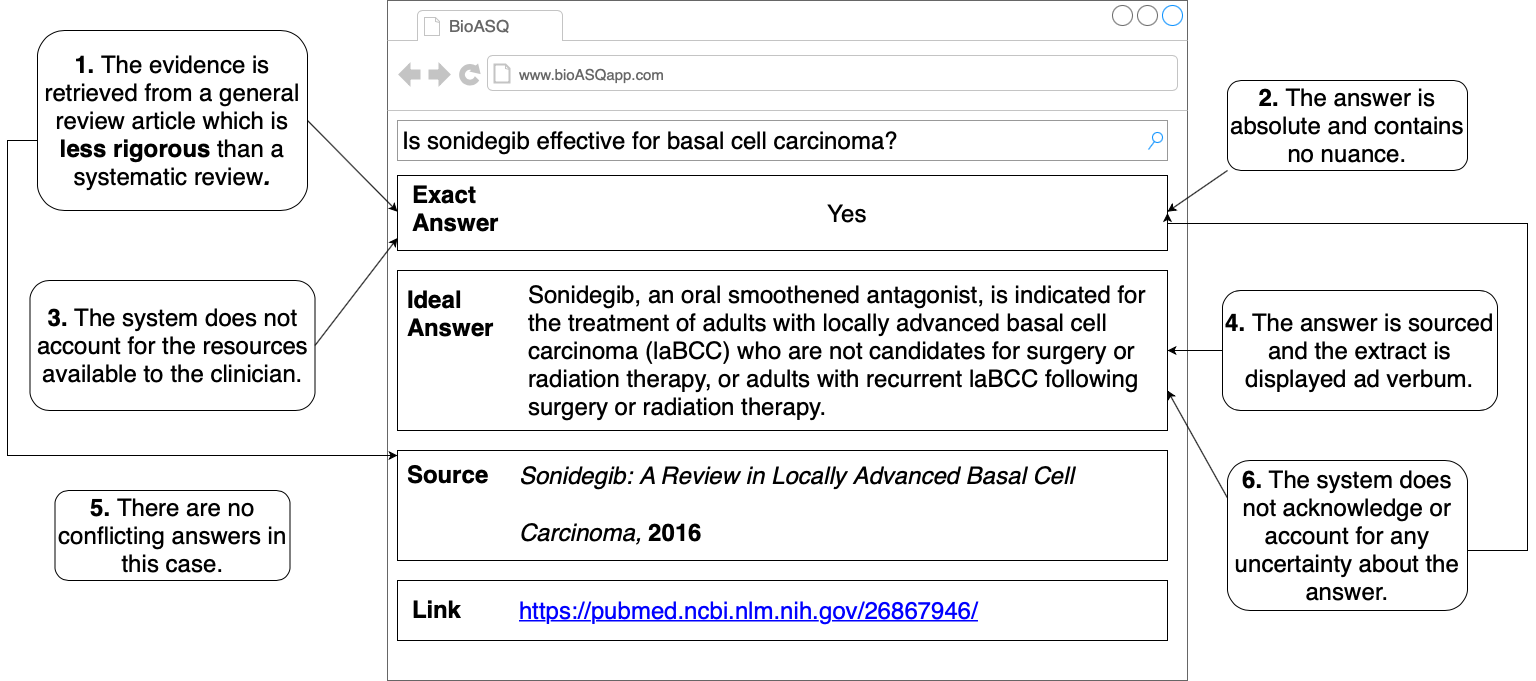}
  \caption{Web interface for QA system developed using BioASQ.}
  \label{fig:BioASQInterface}
\end{figure*}

\begin{figure*}[!ht]
  \centering
  \includegraphics[scale=0.3]{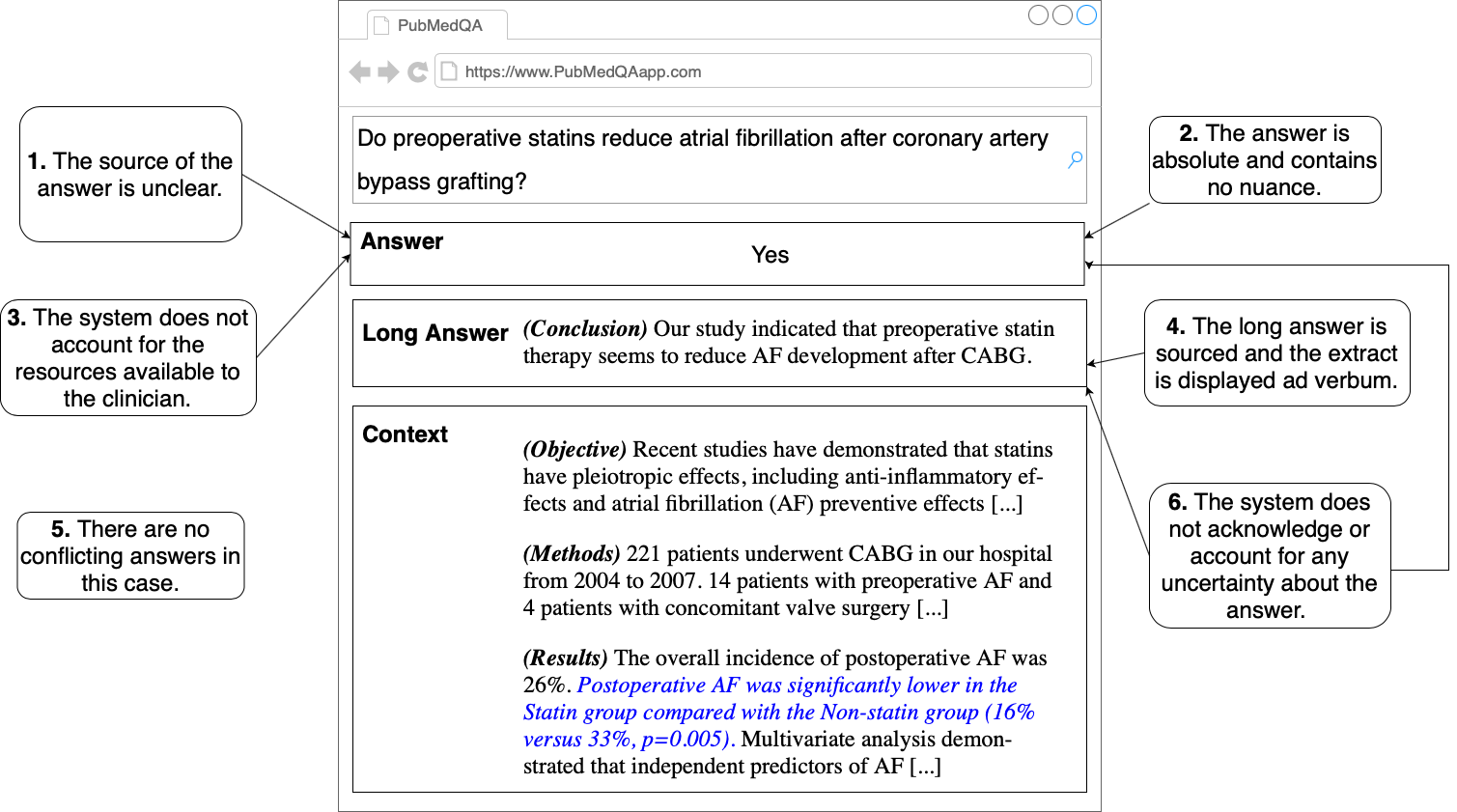}
  \caption{Web interface for QA system developed using PubMedQA.}
  \label{fig:PubMedQAInterface}
\end{figure*}

\begin{figure*}[!ht]
  \centering
  \includegraphics[scale=0.3]{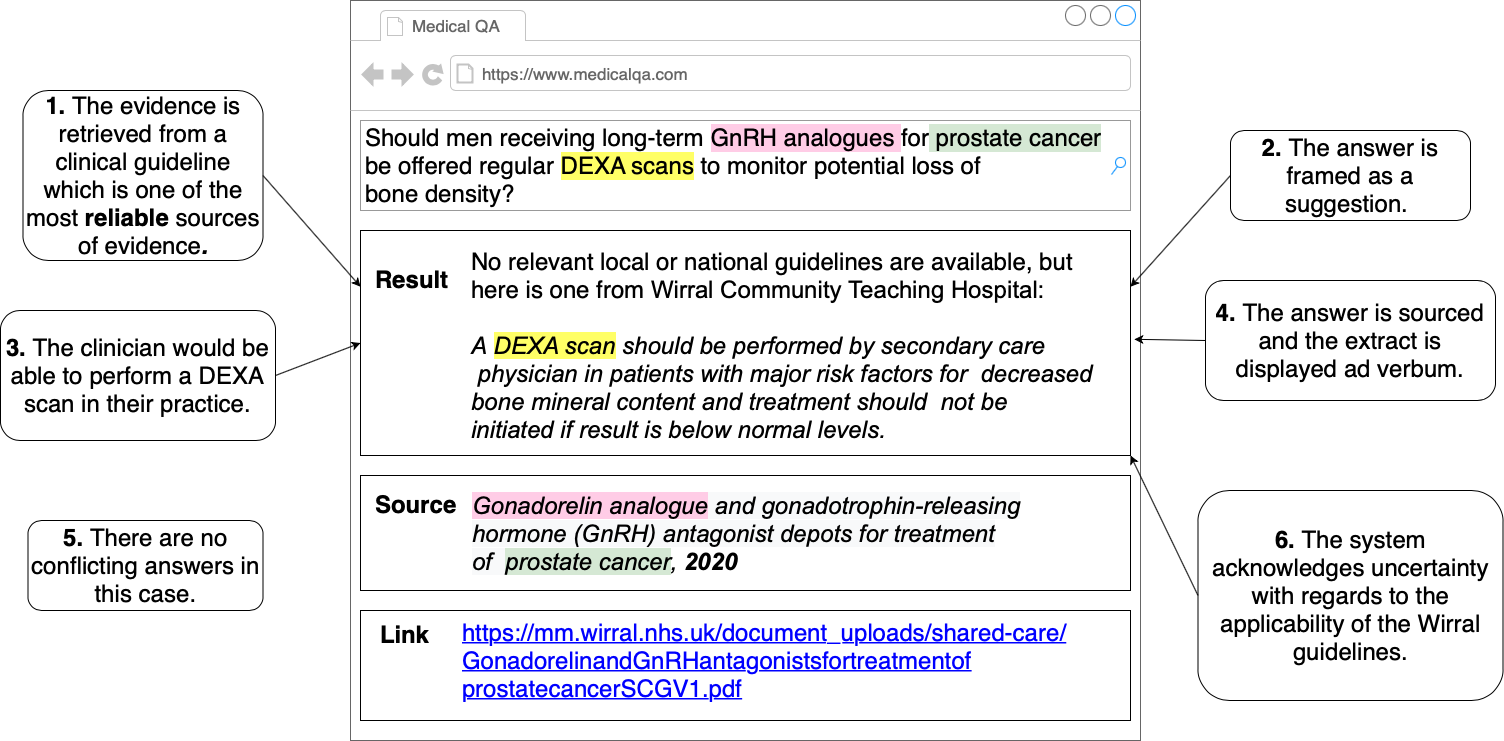}
  \caption{Example of a medical QA system output that meets the criteria in \S\ref{sec:InterpMedicalQA}.  The assumption is that the clinician is based in Nottingham while the most relevant guideline is for Wirral Community Teaching hospital (which is in a different region). The corresponding text spans in the question and response are highlighted with the same color.}
  
  \label{fig:TextualAnswersInteface}
\end{figure*}

\begin{figure*}[!ht]

  \centering
  \includegraphics[scale=0.3]{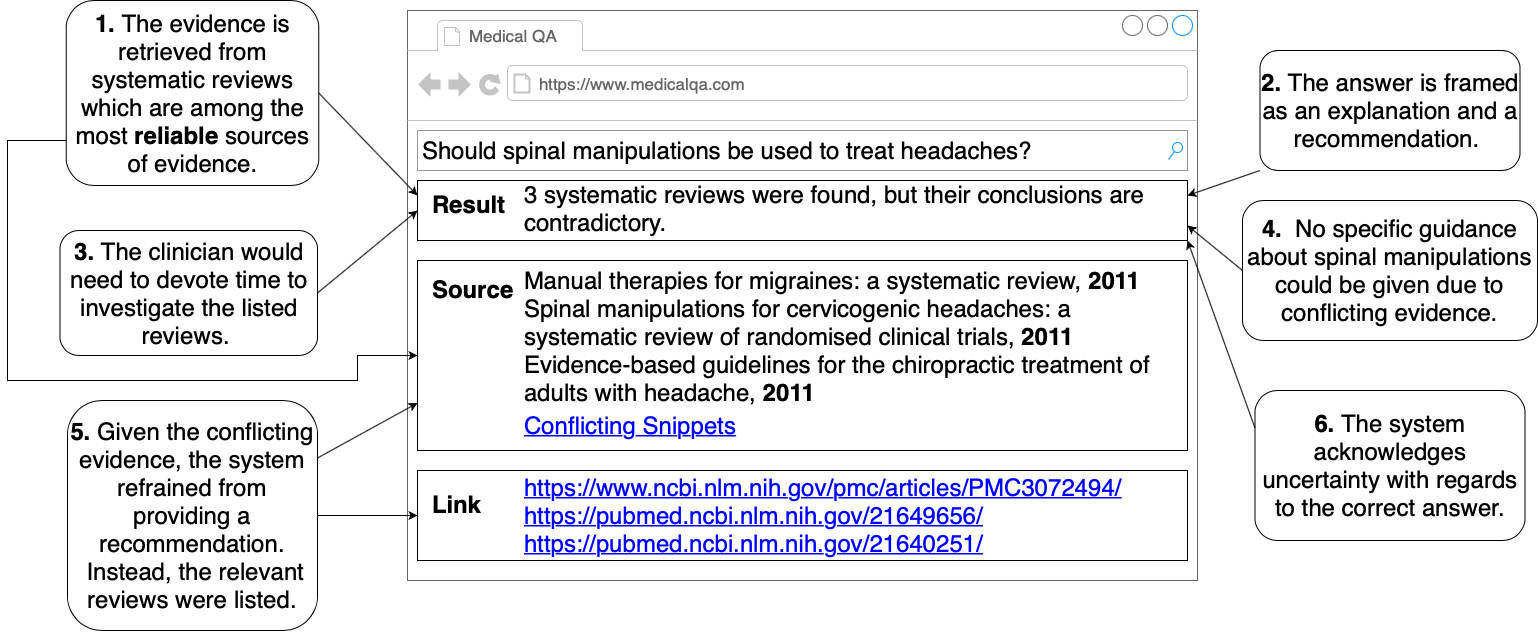}
  \caption{Example of a medical QA system output that handles the conflicting conclusions of 3 systematic reviews.}
  \label{fig:ConflictingEvidenceInterface}
\end{figure*}

\begin{figure}[!ht]

  \centering
  \includegraphics[scale=0.4]{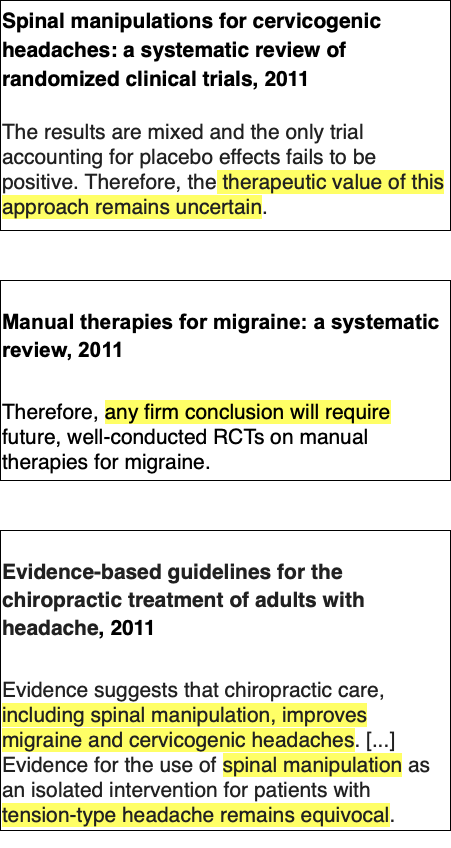}
  \caption{Contradictory source snippets leading to the response presented in figure  \ref{fig:ConflictingEvidenceInterface}.}
  \label{fig:ConflictingSnippets}
\end{figure}

\section{Desiderata for Medical QA}
\label{sec:InterpMedicalQA}




What would be needed for clinicians to trust, and actually act upon answers provided by a QA system?  In our view, the necessary criteria include: Provenance of the evidence and its reliability; Faithfulness of the evidence to the source, and; Transparency with respect to how answers are chosen, and how conflicting evidence is resolved. 
In accordance with these criteria, we suggest the following questions to assess the transparency 
of QA systems:

\begin{enumerate}
    \item \textbf{Do the answers come from reliable sources for health information?} 
    All research articles are not equal, and there exist mature approaches to help clinicians identify the most reliable advice from the health literature. Evidence-Based Medicine is one such framework in which the findings of the most rigorous study designs (typically high quality clinical guidelines, and \emph{systematic reviews} of the primary literature) are preferred to case studies and observational research \cite{alper2016ebhc, sackett1985clinical}. 
    
    More sophisticated approaches (e.g., \emph{risk of bias} assessment tools and the \emph{GRADE} framework; \citealt{higgins2011cochrane, guyatt_grade_2008}) go further by estimating how confident one should be in a research finding, taking into account aspects such as study type, the precision of the statistical results, and whether problems in study design were likely to have led to bias.
    QA systems which take a naive approach to evidence extraction---for example, selecting an answer from an undifferentiated corpus of scientific literature, treating all studies as equally reliable---are likely to be considerably less useful to clinicians. 
    This is particularly true because there is often no definitive ``correct'' answer to a query; an overview of the best available evidence is what is sought. 
    We suggest that QA systems should aim to explicitly use more rigorous, theoretically informed approaches to sorting the literature, mirroring the best current practice of manual question answering and evidence synthesis.
    
    \item \textbf{Does the system provide guidance?}  When searching for answers clinicians are looking for \emph{guidance}, not just information. Guidance consists of recommendations of what to do in various clinical situations, while Boolean or factoid answers appear more absolute. The demand for guidance is reflected by the fact that many questions are of the form ``Should I ...?'' \cite{del_fiol_clinical_2014,ely_taxonomy_2000,papermaster_common_2017}. 
    Therefore, the system could respond with ``study/review X suggests the following action... ''.  This response could encourage the clinician to engage with the guidance and think critically about how to apply it in practice.
    
    
    In the aforementioned example on urinary tract infections (UTI), the NICE\footnote{The National Institute for Health and Care Excellence: the UK national health guideline producer } guideline \cite{niceUTI} recommends Nitrofurantoin under specific conditions: If the estimated glomerular filtration rate (eGFR) $\geq$ 45 ml/minute then 100 mg modified-release twice a day (or if unavailable, 50 mg four times a day) for 3 days.

    \item \textbf{Are the answers useful in the context in which the provider is practicing?}  The usefulness of a QA system could be limited by factors such as drug availability, antibiotic resistance, and local or national funding/resources. Therefore, QA systems should account for the resources that are available to clinicians when providing guidance. In addition, what is deemed as ``best practice'' may vary by location (i.e., region or country).
    
    If a clinician were to consult a QA system on whether ``men receiving long-term GnRH analogues for prostate cancer should be offered regular DEXA scans to monitor potential loss of bone density'', guidelines from Wirral Community Teaching Hospital might be retrieved. The clinician would need to decide whether the guidelines apply to their locality (e.g., Nottingham) where DEXA scans may or may not be readily available.
    
    \item \textbf{Is there sufficient ``rationale'' for the answer provided?} 
    Prior work has shown that users of QA systems prefer answers to consist of paragraph-sized chunks of text as opposed to concise phrases \cite{Lin03whatmakes}. 
    Lengthier ``answers'' provide context, and allow users to ensure that the information in the source text is consistent with the final answer. 
    As answers should be faithful to the source, any generated summaries should probably be extractive rather than abstractive.
    
    
    For example, the answer to ``what dose of flucloxacillin should I prescribe for a 5 year old child?'' could consist of the snippet highlighted in Figure \ref{fig:rationale_extracted_text}.
    However, in cases where the answer is derived from multiple sources it may be necessary to generate a summary.
    
    
    

    \item \textbf{Does the system resolve conflicting evidence appropriately?}  Higher quality information should be prioritized using frameworks for rating the quality of evidence \cite{ebell_strength_2004, guyatt_grade_2008, alper_ebhc_2016}.  If there are conflicts between equally relevant and reliable sources, the system should refrain from providing oversimplified guidance and inform the clinician of the conflicting sources. This could form the basis for further investigation by the clinician.
    
    The query ``Should spinal manipulations be used to treat headaches?'' could return three conflicting systematic reviews: one concluding that they should \cite{bryans_evidence-based_2011} and two others that judge the evidence to be inconclusive \cite{chaibi_manual_2011,posadzki_spinal_2011}.  A QA system should inform the clinician of these contradictions. An ideal system would assess the relative methodological quality of the reviews, and present the most rigorous and reliable first. 
    
    \item \textbf{Does the system handle and communicate uncertainties adequately?}  When providing guidance, the system should communicate any sources of uncertainty. If appropriate, the system should abstain from providing explicit guidance (e.g., where information conflicts or where supporting evidence is either absent or of low quality).
    
    In the case of the regular DEXA scans for men recieving long-term GnRH analogues, the system should communicate its uncertainty on whether the guidelines from Wirral Community Teaching hospital are applicable to the clinician's region.

    Additionally, the question ``Does speech and language therapy help dysarthria after a brain injury?'' could return no relevant studies \cite{sellars_speech_2002}. 
    It is important that the system explain that the question is unanswerable using the available literature. 
\end{enumerate}

%
%
%

There are several research challenges associated with the above criteria. 
Reframing the QA task will require new datasets which include answers (with accompanying rationales) from trusted sources; rankings by evidence quality; locality and patient contextualizing information; and which incorporate real-world conflicting answers and questions which lack answers. 
Quantitative measures would need to be created to assess how well the datasets and systems meet each criterion.

While we expect that an improved system using these criteria might be more trustworthy (and hence potentially help to translate health research more effectively info clinical practice), we note that our criteria need to be empirically tested. 
To achieve this, we need to move beyond dataset evaluation, and consider user-centred design methodology. 
Ultimately, we should aim to improve and evaluate systems through research conducted in real-world clinical practice.

We next review prior work on Biomedical QA with respect to the above criteria. 
We display typical responses of these systems in a hypothetical web interface, and assess how well these responses meet the criteria.

\section{Existing Medical QA Datasets and Systems}
\label{sec:RelatedWork}


\begin{table*}
\centering

 \begin{tabular}{l c c c c c c} 
 \hline
 \thead{QA System/\\ Dataset} & D1 & D2 & D3 & D4 & D5 & D6 \\ [0.5ex] 
 \hline
 BioASQ \cite{Krallinger2020BioASQAC} & \xmark & \xmark  & \xmark & \cmark &  \xmark & \xmark \\ 
PubMedQA \cite{jin_pubmedqa_2019} & \xmark & \xmark & \xmark & \cmark & \xmark & \xmark \\ 
 
 MEDIQA-QA \cite{ben-abacha-etal-2019-overview} & \cmark & \cmark & \xmark & \cmark & \xmark & \xmark  \\ 
 MEDIQA-AnS \cite{savery_question-driven_2020} & \cmark & \cmark & \xmark & \xmark & \xmark & \xmark  \\ 
 LiveQA-Medical \cite{Abacha2017OverviewOT} & \cmark & \cmark & \xmark & \cmark & \xmark & \xmark  \\ 
  MEANS \cite{ben_abacha_means_2015} & \xmark & \xmark & \xmark & \cmark & \xmark & \xmark  \\ 
  AskHERMES \cite{cao_askhermes_2011} & \cmark & \xmark & \xmark & \cmark & \xmark &  \xmark \\ 
  CLINIQA \cite{zahid_cliniqa_2018} & \xmark & \cmark & \xmark & \cmark & \xmark & \xmark \\
  MedQA \cite{10.1016/j.jbi.2007.03.002} & \xmark & \cmark & \xmark & \cmark & \xmark & \xmark \\ [1ex] 
 \hline
\end{tabular}
\caption{Comparision of how well QA systems and datasets meet the desiderata outlined in \S \ref{sec:InterpMedicalQA}.\label{table:QAcomparision}}
\end{table*}

The primary focus of prior medical QA work has been on developing systems that answer the following types of questions: boolean (yes/no), factoid, list (of factoids), and definitional, e.g. \cite{sarrouti_sembionlqa_2020, ben_abacha_means_2015, cao_askhermes_2011, zahid_cliniqa_2018, 10.1016/j.jbi.2007.03.002}. Several datasets have been created to train and evaluate systems that handle the aforementioned question types, including BioASQ \cite{tsatsaronis_overview_2015}, emrQA \cite{pampari_emrqa_2018}, emrKBQA \cite{Raghavan2021emrKBQAAC}, PubMedQA \cite{jin_pubmedqa_2019}, why-QA \cite{fan-2019-annotating}, MEDIQA-QA \cite{ben-abacha-etal-2019-overview},  LiveQA-Medical \cite{Abacha2017OverviewOT} and MEDIQA-AnS \cite{savery_question-driven_2020}. BioASQ, PubMedQA, MEDIQA-QA, MEDIQA-AnS and LiveQA-Medical derive answers from a corpus of biomedical literature, whereas emrQA, emrKBQA and why-QA are based on patient notes within EHRs. As stated above, our focus here is on systems that can answer general questions (independent of individual patients) based on the latest evidence, so we do not discuss emrQA, emrKBQA and why-QA. A comparison of the systems and datasets is provided in Table \ref{table:QAcomparision}.

While BioASQ, MEDIQA-QA, MEDIQA-AnS and LiveQA-Medical are large-scale information retrieval (IR) and question answering (QA) datasets, PubMedQA is designed for ``reading comprehension'' question answering (RCQA) based on scientific abstracts. 
Each question of PubMedQA is accompanied by the abstract containing the answer.

The BioASQ Phase B challenge comprises the following question types \cite{tsatsaronis_overview_2015}:

\begin{itemize}
    \item Exact: ``yes'' or ``no'', e.g., ``Is the protein Papilin secreted?'';
    \item Factoid: named entities, e.g., ``Name synonym of Acrokeratosis paraneoplastica.'';
    \item List: list of named entities, e.g., ``Which miRNAs could be used as potential biomarkers for epithelial ovarian cancer?'';
    \item Ideal: paragraph-sized summaries (text spans), e.g., ``What is the effect of TRH on myocardial contractility?''
\end{itemize}

While BioASQ has been instrumental to the progress of the field \cite{nentidis-etal-2017-results,nentidis-etal-2018-results,Nentidis_2020,Krallinger2020BioASQAC}, it 
satisfies only one of the criteria we have enumerated above, namely 4. 
Figure \ref{fig:BioASQInterface} shows the expected output of a system developed using BioASQ.  In this example, the extract is provided 
verbatim (criterion 4). 

However, the answer is sourced from a general review; these reviews are 
less reliable than guidelines or systematic reviews (criterion 1).  Furthermore, the system outputs absolute answers rather than guidance (criterion 2) which limits their usefulness to clinicians.  A more suitable answer would be ``the following guidance is provided in X...''.  It is unclear what resources are available to the clinician and the BioASQ dataset does not account for this (criterion 3). 
There is no contradictory evidence in the example and BioASQ has been preprocessed to ensure there are no conflicting papers (criterion 5). 
Unless the trained model is acting on a curated knowledge base, it would not be robust to conflicts. 
Finally, the absolute nature of the answer does not allow the system to recognise and account for uncertainty (criterion 6).

In contrast to BioASQ, PubMedQA provides answers to only Boolean (yes/no) questions, e.g. ``Do preoperative statins reduce atrial fibrillation after coronary artery bypass grafting?''. 
Accompanying these responses is a ``long answer'', supplied in the form of the conclusions of the source abstracts. 
As per Figure \ref{fig:PubMedQAInterface}, the outputs of systems trained on PubMedQA can only satisfy criterion 4. 
The conclusion is given 
verbatim to support the short answer. 
Nevertheless, the source of the answer is not specified (criterion 1), the answer is absolute (criterion 2) and it does not account for any uncertainty (criterion 6). 
Systems developed using PubMedQA cannot ensure that the answer is useful to the clinician (criterion 3). 
Given the task is framed as ``reading comprehension'', there is only one abstract per question. 
This prevents systems from being trained to handle conflicts (criterion 5).

MEDIQA-QA is a consumer QA dataset whose answers consist of exact snippets from MedlinePlus.  Consumer questions are more focused on general information, symptom or  person/organization questions \cite{roberts_interactive_2016}.  The answers that are required by consumers are less complex and more easily understandable than those given to clinicians \cite{savery_question-driven_2020}.  This has motivated the development of MEDIQA-AnS which summarises the answers of MEDIQA-QA. As shown in Figures \ref{fig:MEDIQA-QA} and \ref{fig:MEDIQA-AnS}, MEDIQA-QA satisfies desiderata 1,2 and 4 while MEDIQA-AnS satisfies only 4.

Although the LiveQA-Medical dataset uses the same answers and sources as MEDIQA-QA and MEDIQA-Ans, it differs by providing answers to each subquestion of the query. Additionally, verbatim extracts of MedlinePlus are used in the responses (Figure \ref{fig:LiveQA}). Hence criteria 1, 2 and 4 are fulfilled.

MEANS returns only an extract of the original source, without any contextualizing information (Figure \ref{fig:MEANS}), i.e. the provenance of the answer. Therefore, only condition 4 is satisfied.

On the other hand, AskHERMES provides a list of answers which are labelled with topics from the question (Figure \ref{fig:AskHERMES}). The extracts shown are from the original sources and are accompanied by links, authors, and dates. Thus, AskHERMES satisfies desiderata 1 and 4.

CLINIQA responds to queries with original abstracts that are accompanied with the PMID and the title of the source paper (Figure \ref{fig:CLINIQA}).  However, the results are not rank according to reliability, so only criteria 2 and 4 are met.

Finally, MedQA's answers comprise sourced extracts from Medline and Google:Definition (Figure \ref{fig:MedQA}).  Answers are not ranked according to reliability, so the system only satisfies criteria 2 and 4. 

None of the aformentioned datasets or systems address conflicts (criterion 5) or communicate uncertainty to clinicians (criterion 6). 
What might QA systems that satisfy all desiderata look like?

\section{Presentation of Answers}
\label{sec:Justifications}

We have seen that systems trained on BioASQ and PubMedQA do not satisfy all the criteria defined in \S\ref{sec:InterpMedicalQA}. 
In this section we present 
illustrative outputs of hypothetical systems that meet the full set of criteria we have put forth. 

Figure \ref{fig:TextualAnswersInteface} presents an example output which satisfies the criteria but where no conflicts occur (criterion 5).  
The answer is sourced from a systematic review (criterion 1) and is in the form of guidance (criterion 2).  
While the guidance is actionable given the resources available (criterion 3) and the source extract is reproduced directly (criterion 4), the uncertainty in the answer is acknowledged (criterion 6) by stating the absence of relevant local and national guidelines. 
The corresponding words and phrases in the question, answer and title used to extract the text snippet are highlighted.

A demonstration of how conflicting evidence could be addressed is provided in Figure \ref{fig:ConflictingEvidenceInterface}. 
In this scenario, the question ``Should spinal manipulations be used to treat headaches?'' returned three contradictory systematic reviews \cite{bryans_evidence-based_2011,chaibi_manual_2011,posadzki_spinal_2011} (criterion 1).  Therefore, the system refrains from providing explicit guidance (criterion 6) and instead provides the clinician with the names and links of conflicting reviews (criterion 5). In addition, the clinician is able to investigate the contradictory snippets further by clicking on "Conflicting Snippets" which would show the snippets in Figure \ref{fig:ConflictingSnippets}.  
Criterion 4 is inapplicable in this case as no answer was retrieved from the documents.

One promising direction which may permit improved handling of contradictory evidence involves use of argumentation-based logic to ``reason'' about multiple potentially conflicting inputs \cite{10.5555/3306127.3332107, Cyras2018ArgumentationFE}, perhaps after explicitly inferring the reported findings concerning treatment efficacies \cite{lehman2019,nye2020understanding}. An alternative (more audacious) direction would be to generate comparative summaries for clinicians that 
compose narrative summaries of the evidence on a given topic from primary sources, including discussion of conflicting evidence \cite{Wallace2020GeneratingN,Shah2021NutribulletsHM}.

Developing and assessing systems according to the criteria outlined in \S\ref{sec:InterpMedicalQA} would ensure the output is useful, actionable and reliable to clinicians. 
It would additionally improve the accountability of both the clinician and the system as the form of the output would be conducive to debugging and root cause analysis.



\section{Conclusions}

We have introduced criteria for assessing the transparency of medical question answering systems. These have been guided by the following question: What would be needed for clinicians to trust, and act upon answers from a QA system? In part we have argued that these systems should be explicitly informed by principles of EBM. 
The adequacy of existing medical systems and datasets, including BioASQ, PubMedQA, MEDIQA-QA, MEDIQA-AnS, LiveQA-Medical, MEANS, AskHERMES, CLINIQA and MedQA, was assessed using the transparency criteria that we proposed.  We found that they met some, but not all, of the conditions.

We presented hypothetical examples of system outputs that satisfy all of the criteria and explained how they could be useful to clinicians. 
These included conflicts between sources of similar reliability. 
In these cases, the best course of action was to refrain from giving guidance and instead return the sources to the clinicians for further examination.  The examples could form the basis of new datasets and systems that provide actionable answers to clinicians.



We believe that these avenues of investigation would assist with the \emph{deployment} of medical QA systems, ultimately furthering the practice of EBM.

\section{Acknowledgements}

This work was supported in part by the National Institutes of Health (NIH), grant R01-LM012086. 

GK holds a doctoral studentship co-sponsored by Metadvice and the Guy's and St Thomas' Biomedical Research Centre.

\bibliography{anthology,custom}
\bibliographystyle{acl_natbib}

\appendix

\section{Additional figures of QA interfaces}

\begin{figure*}[!ht]
  \centering
  \includegraphics[scale=0.25]{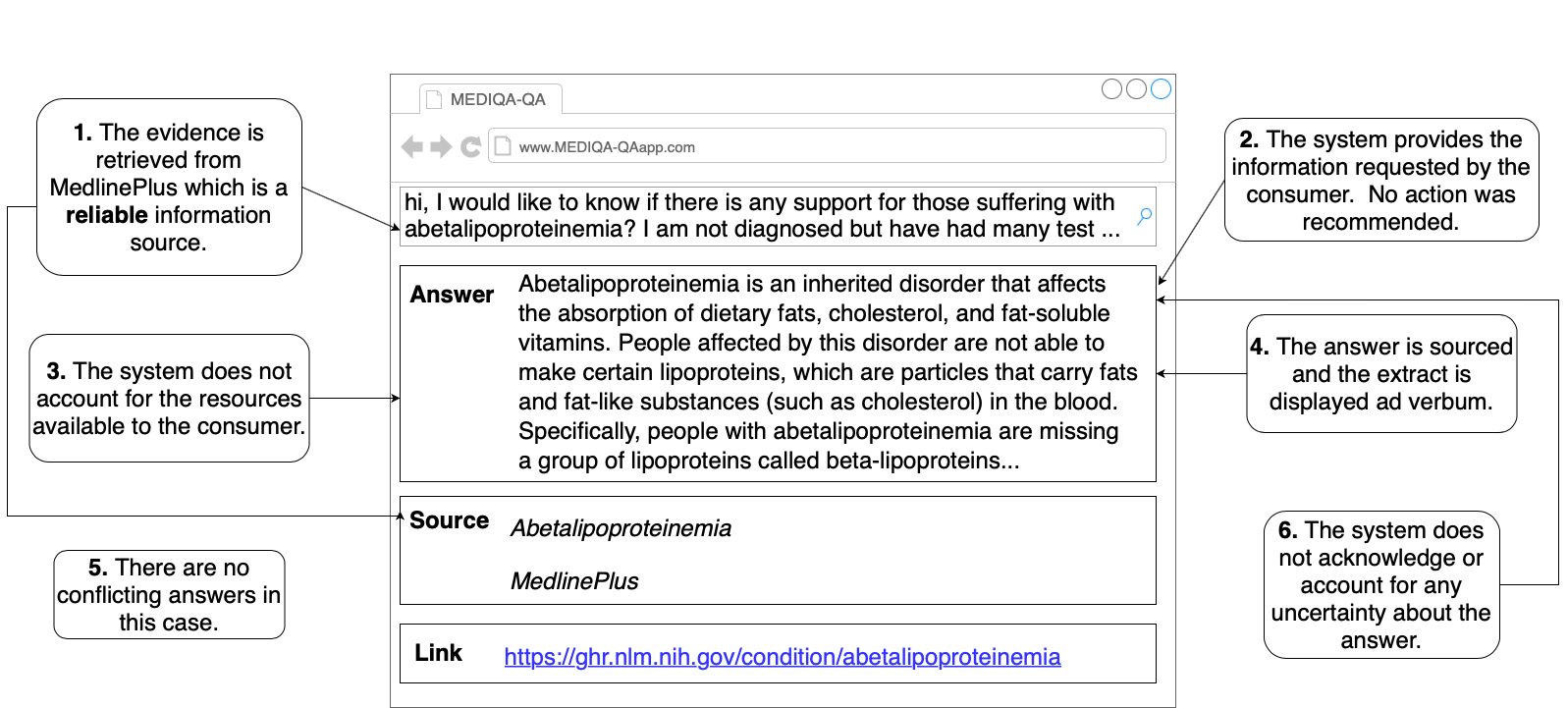}
  \caption{Web interface for QA system developed using MEDIQA-QA.}
  \label{fig:MEDIQA-QA}
\end{figure*}

\begin{figure*}[!ht]
  \centering
  \includegraphics[scale=0.25]{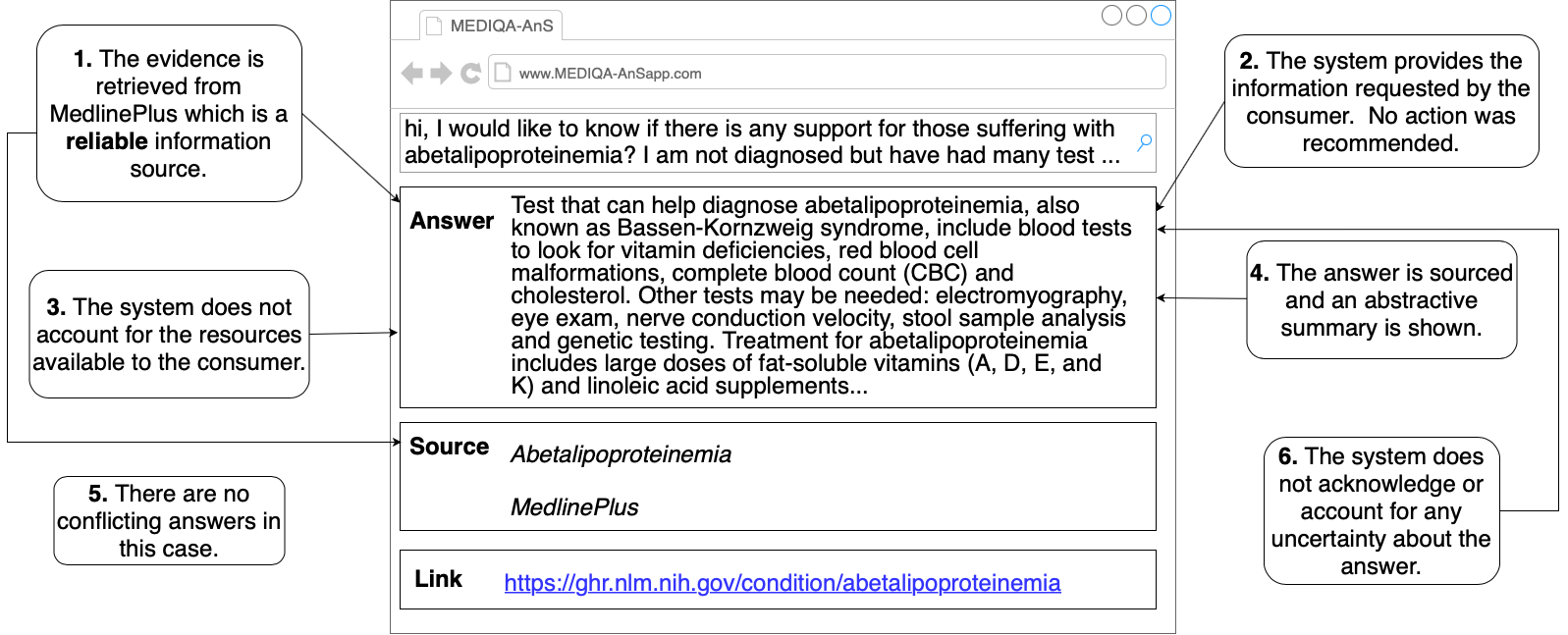}
  \caption{Web interface for QA system developed using MEDIQA-AnS.}
  \label{fig:MEDIQA-AnS}
\end{figure*}

\begin{figure*}[!ht]
  \centering
  \includegraphics[scale=0.25]{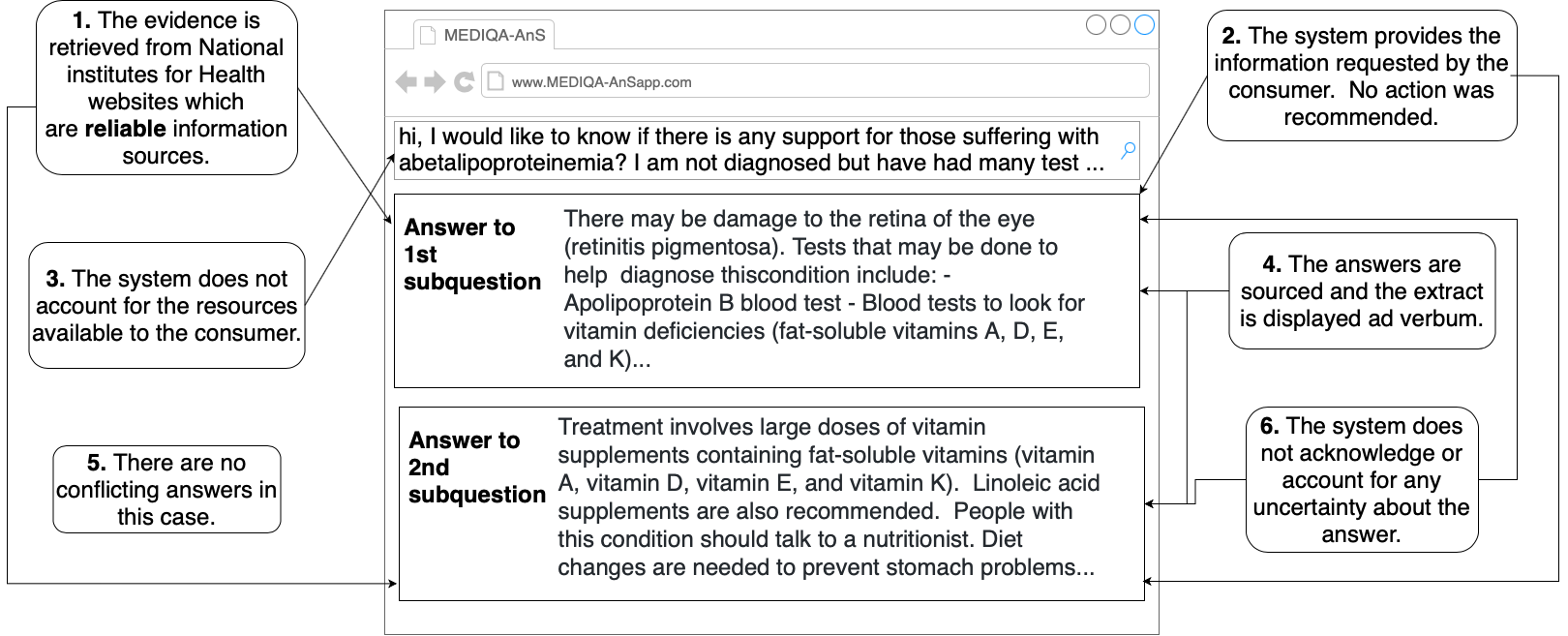}
  \caption{Web interface for QA system developed using LiveQA-Medical.}
  \label{fig:LiveQA}
\end{figure*}

\begin{figure*}[!ht]
  \centering
  \includegraphics[scale=0.25]{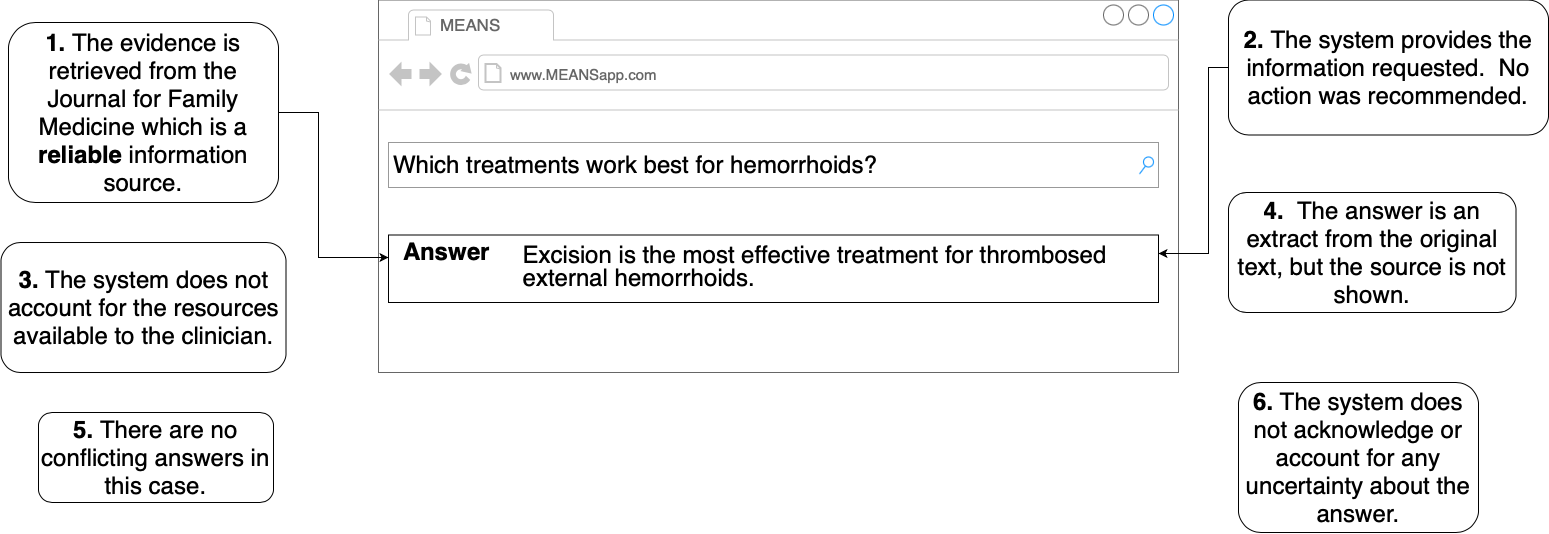}
  \caption{Web interface for MEANS.}
  \label{fig:MEANS}
\end{figure*}

\begin{figure*}[!ht]
  \centering
  \includegraphics[scale=0.25]{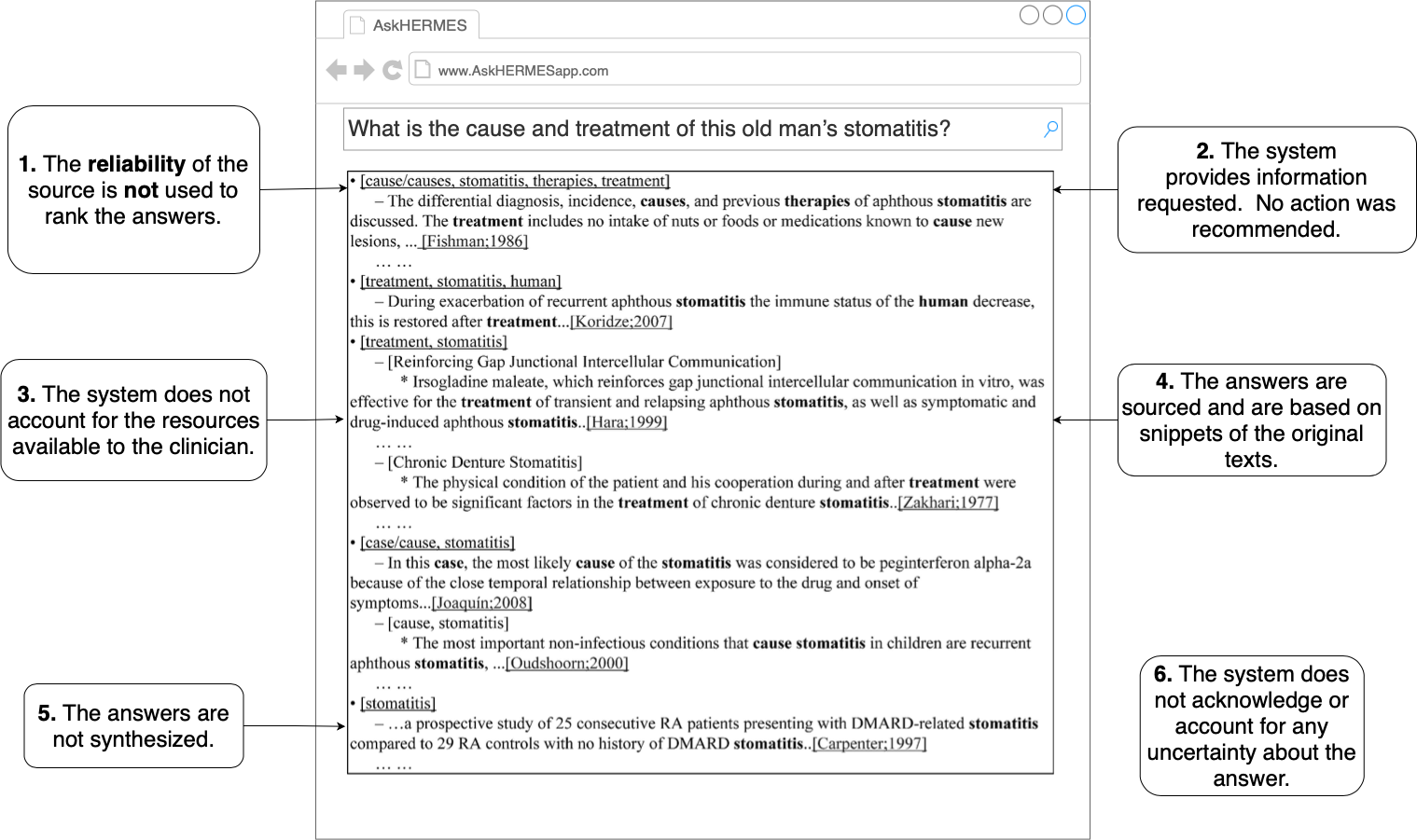}
  \caption{Web interface for AskHERMES.}
  \label{fig:AskHERMES}
\end{figure*}

\begin{figure*}[!ht]
  \centering
  \includegraphics[scale=0.25]{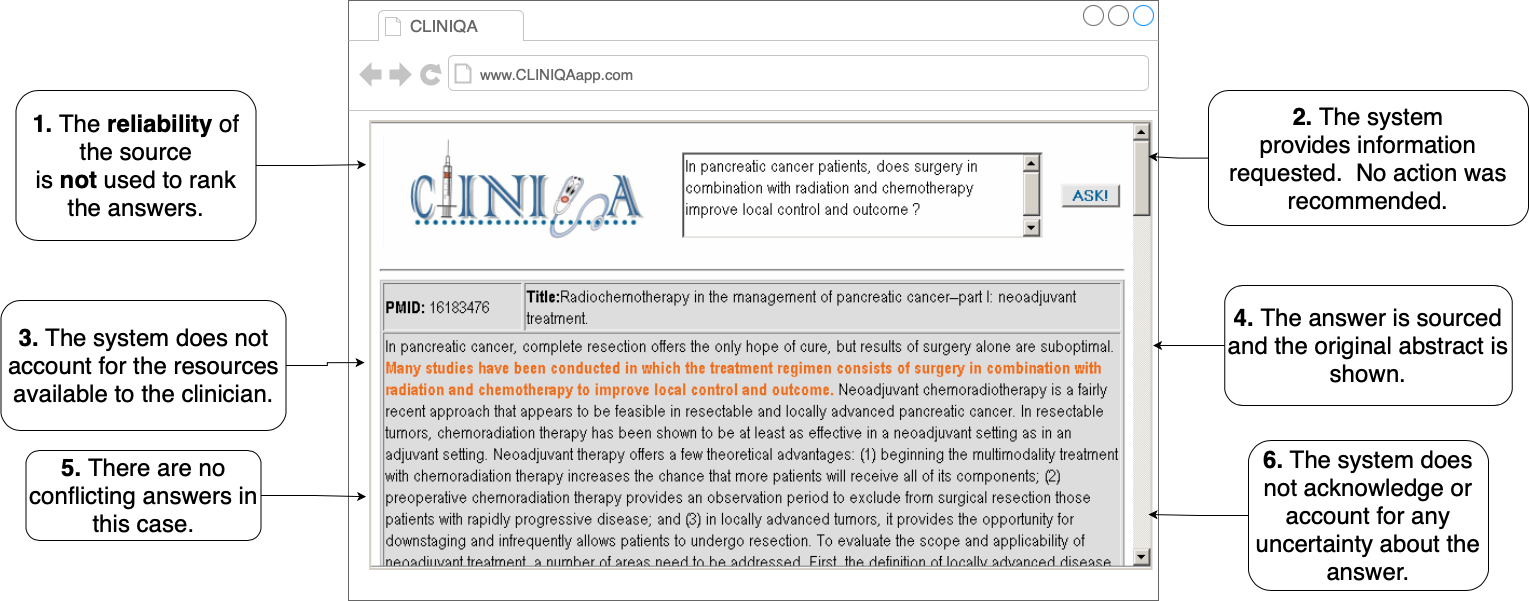}
  \caption{Web interface for CLINIQA which includes figure 5 from \cite{zahid_cliniqa_2018}.}
  \label{fig:CLINIQA}
\end{figure*}

\begin{figure*}[!ht]
  \centering
  \includegraphics[scale=0.25]{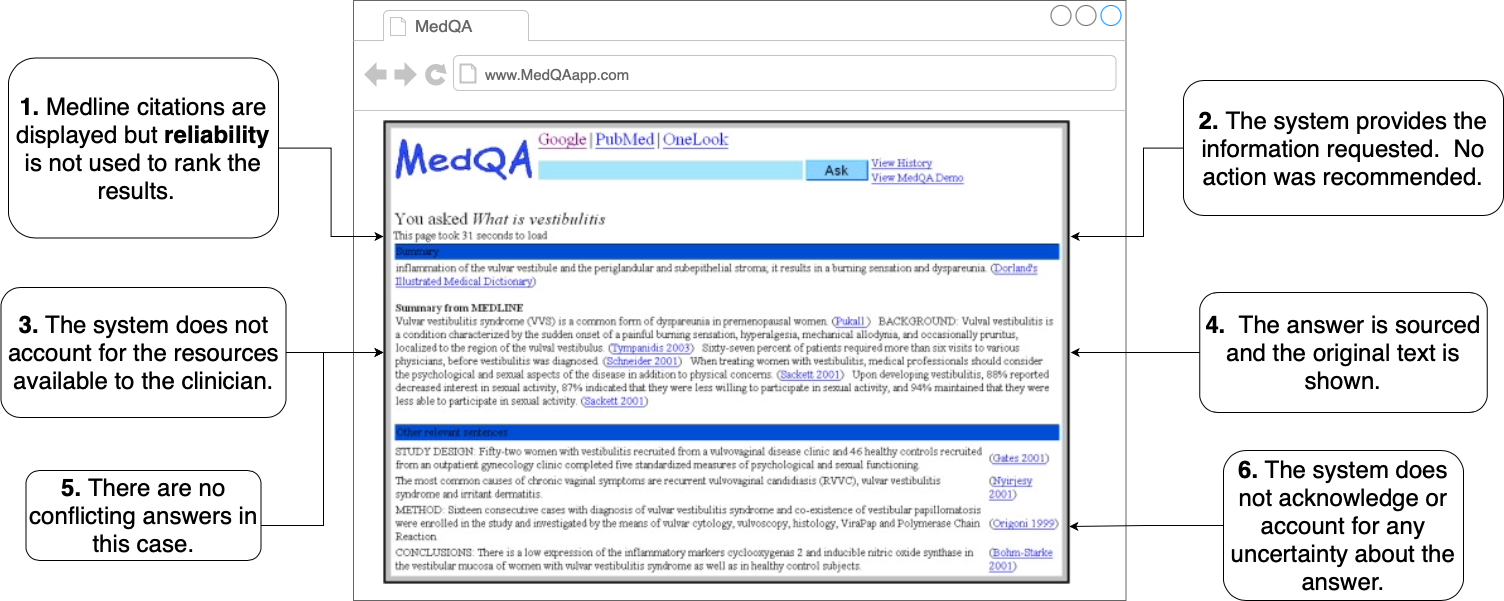}
  \caption{Web interface for MedQA which includes figure 3 from \cite{10.1016/j.jbi.2007.03.002}.}
  \label{fig:MedQA}
\end{figure*}


\end{document}